\RequirePackage{fix-cm}
\documentclass[]{dataflow}

\usepackage[toc,page,header]{appendix}

\usepackage{minitoc}
\usepackage{multirow}
\usepackage{multicol}
\usepackage{booktabs}
\usepackage{colortbl}
\usepackage[table]{xcolor}
\usepackage{makecell}
\usepackage{graphicx}
\usepackage{amssymb}
\usepackage{pifont}
\usepackage{fontawesome5}
\usepackage{float}
\usepackage{adjustbox}

\title{DataFlex-RL: An Evaluation Platform for RLVR Data Policies}

\author[*, 1, 4]{Hao Liang}
\author[*, 2, 4]{Mingrui Chen}
\author[*, 1]{Hengyi Feng}
\author[*, 1]{Meiyi Qiang}
\author[\ddagger, 1, 3, 4]{Wentao Zhang}

\affiliation[]{$^{1}$Peking University, $^{2}$UCAS, $^{3}$Institute for Advanced Algorithms Research, Shanghai, $^{4}$Zhongguancun Academy}

\contribution[*]{Equal Contribution}
\contribution[\ddagger]{Corresponding author}

\abstract{
Data policies for reinforcement learning with verifiable rewards (RLVR) change which rollouts are used, how strongly they are weighted, or which domains supply the next batch. We introduce \textsc{DataFlex-RL}, an evaluation platform for comparing these choices under the same GRPO recipe. Our primary experiment evaluates 13 configurations with 12 matched seeds on Qwen2.5-7B-base and 12 math, logic, and science benchmarks. Uniform GRPO improves the domain-balanced average accuracy (Overall) by $7.76$ points over the untrained checkpoint. None of the eight selection or reweighting methods has a paired 95\% confidence interval excluding zero relative to uniform sampling, and none of the three adaptive mixtures improves over a fixed equal mixture at that precision. A corrected 12-seed Llama-3.1-8B-base extension puts the additional methods on the same score scale as the original controls, without producing a common winner in their observed means. We also quantify evaluation sensitivity by recomputing nine Qwen2.5-7B-Instruct runs with a math-heavy six-benchmark summary---five math benchmarks plus GPQA-Diamond, with no logic benchmark---and with the domain-balanced 12-benchmark summary. Their rankings are negatively correlated ($\rho=-0.33$), while summaries retaining all 12 benchmarks largely agree. Across the controlled settings studied here, changing the data policy measurably changes the training process but does not yield a reproducible improvement over uniform training.
}

\date{\today}

\def\emailicon{\raisebox{-1.5pt}{\includegraphics[height=1.05em]{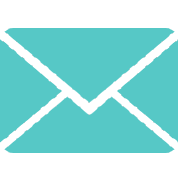}}}
\def\githubicon{\raisebox{-1.5pt}{\includegraphics[height=1.05em]{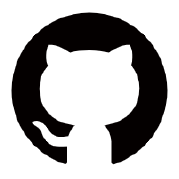}}}

\newcommand{\sourcelink}{https://github.com/haolpku/DataFlex-RL}

\checkdata[ \emailicon \hspace{0.3em} Correspondence ]{\email{wentao.zhang@pku.edu.cn}}
\checkdata[ \githubicon \hspace{0.3em} Source Code ]{ \url{\sourcelink} }
\checkdata[ \faFile \hspace{0.57em} Results \& Raw Numbers ]{ \url{https://github.com/haolpku/DataFlex-RL/tree/main/results} }
\checkdata[ \faFile \hspace{0.57em} Codebase Documentation ]{ \url{https://haolpku.github.io/DataFlex-RL-Doc/} }

\begin{document}
\maketitle

\renewcommand{\thefootnote}{\fnsymbol{footnote}}
\setcounter{footnote}{0}
\renewcommand{\thefootnote}{\arabic{footnote}}
\pagestyle{fancy}
\fancyhf{}
\fancyhead[L]{DataFlex-RL Technical Report}
\fancyhead[R]{\thepage}

\newpage
\tableofcontents
\newpage

\section{Introduction}
\label{sec:intro}

Reinforcement learning with verifiable rewards (RLVR) is widely used to post-train reasoning models. In Group Relative Policy Optimization (GRPO)~\cite{shao2024deepseekmath}, each update samples prompts, generates multiple responses, scores them with a verifier, and constructs group-relative advantages. This makes the training distribution an online choice: a data policy decides which prompts receive rollout compute, how domains are mixed, which groups survive filtering, and how strongly their responses contribute to the update.

Prompts that are solved by every rollout or by none of them yield degenerate groups, while prompts near the policy's decision boundary may provide informative comparisons. This observation has produced a diverse set of interventions: solve-rate filtering in DAPO~\cite{yu2025dapo}, high-variance down-sampling in PODS~\cite{xu2025pods}, probability-based Advantage Reweighting~\cite{yang2025lowprob}, advantage weighting inspired by prioritized experience replay~\cite{schaul2016prioritized}, and adaptive domain curricula such as DUMP~\cite{wang2025dump} and teacher-student curriculum learning~\cite{matiisen2017teacher}. These methods alter different parts of the training loop, but all try to concentrate rollout or optimization effort on data judged more useful.

The appeal of these methods rests on a broader premise: directing training toward more useful data should produce gains that survive reasonable changes in model and training condition. Existing evidence does not yet establish that premise. Data-processing methods are often introduced as one component of a larger RLVR recipe, alongside changes to the base model, prompt template, verifier, rollout budget, and optimization settings. Even methods driven by the same signal may use it differently: advantage magnitude can define either a continuous loss weight or a hard selection rule. A reported gain can therefore reflect the signal, the intervention, or the surrounding training stack.

On-policy evaluation adds two further complications. The score attached to a prompt changes as the policy learns, so the data utility signal is both noisy and endogenous to training. At the same time, training-seed variation and measurement noise on small reasoning benchmarks can match or exceed the reported gap between methods. The evaluation summary can introduce another choice: omitting a domain or weighting benchmarks differently may change which method appears best. A training--evaluation format mismatch can also leave optimization traces apparently normal while making benchmark scores invalid. A useful comparison must control the training recipe, cover the intended evaluation domains, align the verifier and evaluation formats, and preserve matched seeds from training through final evaluation.

We therefore ask whether the tested RLVR data policies provide reproducible gains over uniform sampling when the surrounding GRPO recipe is held fixed. The evaluation begins with a complete 12-seed comparison on Qwen2.5-7B-base, where GRPO itself produces a large gain. This experiment tests all selection, reweighting, and mixture configurations in a setting with substantial training headroom. We then extend the comparison to Llama base models and several Qwen base-model sizes, and test whether the conclusion depends on evaluation coverage. Separate controls examine whether selection benefits come from the targeting signal, from changing the number of tokens used for optimization, or from intervention strength.

\textsc{DataFlex-RL} is an evaluation platform built around this comparison. It separates what a policy changes from the signal it uses and transfers the selection--reweighting--mixture abstraction of \textsc{DataFlex}~\cite{liang2026dataflex} from supervised LLM training to on-policy RL. Selection decides which generated responses enter the update, reweighting changes their continuous loss contribution, and mixture adaptation changes which domains supply future prompts. Solve rate, reward, advantage, and token probability are then recorded as signals rather than treated as method categories. This distinction matters experimentally: two methods may use similar scores while changing different quantities, including the effective number of update tokens. The platform pairs this taxonomy with a standardized math/logic/science corpus, fixed GRPO training, domain-specific evaluation harnesses, and a calibration smoke test for training--evaluation alignment.

\begin{figure}[t]
    \centering
    \includegraphics[width=\linewidth]{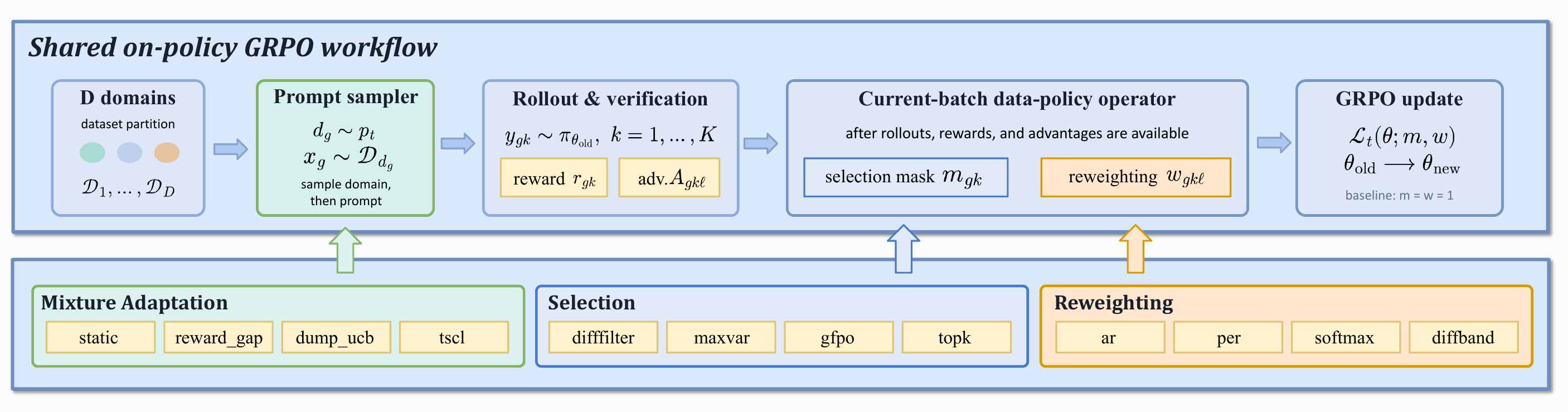}
    \caption{The shared GRPO workflow and the intervention point of each data-policy family. Mixture adaptation changes the domains supplying future prompts; selection and reweighting act after rollout and verification by modifying the current update. Uniform sampling with unit mask and weight is the common baseline.}
    \label{fig:overview}
\end{figure}

The experimental matrix contains 591 runs, including a 156-run primary block that compares all 13 Qwen2.5-7B-base configurations over 12 seeds. For run accounting, the release consists of the original 171-run breadth grid, 108 runs that extend three four-method comparisons from 3 to 12 seeds, 33 mechanism and strength controls, 108 runs that complete the Qwen-base primary matrix, and 171 base-model scale and family runs from Revision Plan~6. Runs are linked to their configurations, training logs, and 12-benchmark records (Figure~\ref{fig:overview}).

The paper has three main empirical findings. First, on Qwen-base, none of the eight selection or reweighting methods has a paired interval excluding zero relative to uniform sampling, and none of the three adaptive mixtures has one relative to the fixed equal mixture; their mean spreads, $0.97$ and $0.61$ point, are small beside the $7.76$-point gain from GRPO itself. Second, the corrected Llama extension finds no common winner across the expanded set of methods. Third, benchmark coverage changes the apparent winner, whereas alternative summaries retaining all 12 benchmarks largely agree. The broader base-model scale and family sweep tests how widely the primary result extends.

\paragraph{Contributions.} We make three contributions:

\begin{enumerate}
    \item \textbf{No clear advantage over uniform sampling in the tested setting.} We evaluate 13 selection, reweighting, and mixture configurations with 12 matched seeds on Qwen2.5-7B-base. Uniform GRPO improves by $7.76$ points, but no selection or reweighting method shows a clear improvement over uniform sampling, and no adaptive mixture improves over a fixed equal mixture at the measured precision (Section~\ref{sec:analysis}).
    \item \textbf{A direct measurement of evaluation sensitivity.} On the same nine Qwen2.5-7B-Instruct configurations, a math-heavy six-benchmark summary (Math-Heavy-6; five math sets plus GPQA-Diamond) and the domain-balanced 12-benchmark summary have negatively correlated rankings ($\rho=-0.33$), with method spread changing from $0.90$ under Math-Heavy-6 to $3.31$ under DB-12. Summaries that retain all 12 benchmarks largely agree, showing that omitting the logic domain can change the apparent winner (Section~\ref{sec:experiments}).
    \item \textbf{A platform for fair comparisons.} \textsc{DataFlex-RL} isolates the data policy from the surrounding GRPO recipe through a shared rollout, verification, optimization, and evaluation stack, matched seeds, and a common 12-benchmark score format. The intervention taxonomy and shared driver let future policies be compared under the same protocol without rebuilding the pipeline (Section~\ref{sec:pipeline}).
\end{enumerate}

\section{Related Work}
\label{sec:related}

\subsection{Data Processing for RLVR}

\paragraph{RL with verifiable rewards.} DeepSeekMath introduced GRPO, which computes group-relative advantages from multiple responses to the same prompt~\cite{shao2024deepseekmath}. DeepSeek-R1 demonstrated strong mathematical and logical reasoning from outcome-verifiable RL~\cite{guo2025deepseekr1}, while Qwen2.5-Math developed a self-improvement pipeline spanning data generation, reward modeling, and RL~\cite{yang2024qwen25math}. DAPO further opened a large-scale RLVR recipe~\cite{yu2025dapo}. These systems evaluate complete training stacks; \textsc{DataFlex-RL} isolates their data-processing component.

\paragraph{Sample selection and reweighting.} DAPO removes groups whose responses are uniformly correct or incorrect~\cite{yu2025dapo}. PODS retains a within-group subset that maximizes reward variance~\cite{xu2025pods}, while GFPO filters responses by length or reward per token~\cite{gfpo2025}. Advantage Reweighting dampens the contribution of low-probability tokens~\cite{yang2025lowprob}; prioritized experience replay instead emphasizes high-surprise transitions~\cite{schaul2016prioritized}. These methods can use related signals while making categorically different interventions: selection sets a response's contribution to zero, whereas reweighting changes it continuously. We compare both under matched models, data, rollout budgets, and seeds.

\subsection{Data-Centric Training}

\paragraph{Offline selection and mixture optimization.} DSIR resamples examples toward a target distribution~\cite{xie2023dsir}, while LESS estimates optimizer-aware influence for instruction tuning~\cite{xia2024less}; pruning and curation can also change pretraining scaling behavior~\cite{sorscher2022beyond,li2024datacomp}. At the domain level, DoReMi derives mixture weights with a proxy model~\cite{xie2023doremi}, and RegMix predicts mixtures from many small proxy runs~\cite{liu2024regmix}. These policies are fixed before the main run and do not respond to on-policy rewards.

\paragraph{Online curricula and unified systems.} Skill-It adapts sampling over prerequisite skills~\cite{chen2023skillit}; teacher-student curriculum learning~\cite{matiisen2017teacher} and DUMP~\cite{wang2025dump} update sampling from observed progress. \textsc{DataFlex} unifies selection, mixture optimization, and reweighting for supervised LLM training~\cite{liang2026dataflex}. \textsc{DataFlex-RL} brings this abstraction to noisy, policy-dependent RL interventions and evaluates them under a common protocol.

\subsection{Evaluation and Reproducibility}

\paragraph{Evaluation infrastructure.} The LM Evaluation Harness~\cite{gao2021lmeval}, HELM~\cite{liang2022helm}, and OpenCompass~\cite{cao2026opencompass} standardize prompts and scoring. Reasoning resources add task-specific protocols: Qwen2.5-Math covers MATH and GSM8K~\cite{yang2024qwen25math,hendrycks2021measuring,cobbe2021training}, while GPQA~\cite{rein2023gpqa}, MMLU-Pro~\cite{wang2024mmlupro}, and ZebraLogic~\cite{lin2025zebralogic} probe science and logic. We retain these harnesses but compare a matched grid of training interventions rather than unrelated fixed checkpoints.

\paragraph{Reproducible empirical RL.} Task choice and evaluation design can change benchmark conclusions~\cite{dehghani2021benchmark}; in RL, implementation details and seeds can reverse them~\cite{henderson2018deep}. Reliable comparisons therefore need intervals and robust aggregation~\cite{agarwal2021statistical}, with explicit treatment of variance, baselines, and tuning~\cite{patterson2024empirical}. Reproducibility practice also emphasizes code, data, and complete procedures~\cite{pineau2021reproducibility}. Our platform supplies aligned evaluation, matched budgets and seeds, intervals, and run-level records.

\section{Three Data-Policy Interventions}
\label{sec:taxonomy}

RLVR data policies differ first in the part of training they change. \emph{Selection} removes generated responses or prompt groups from the current update. \emph{Reweighting} keeps those responses but changes their loss contribution. \emph{Mixture adaptation} changes the domain distribution used for subsequent prompt sampling. We describe these intervention points before discussing the signals that drive them.

\subsection{Selection, reweighting, and mixture adaptation}

The baseline training step is simple: it samples math, logic, and science prompts in equal proportions, generates $K=5$ responses for each prompt, verifies their rewards, and applies the standard GRPO loss to every valid response token. Each policy family below changes one part of this process.

\paragraph{Selection: choose which responses enter the update.}
Selection is applied after the responses and rewards have been generated. Let $m_{gk}\in\{0,1\}$ indicate whether response $k$ to prompt $g$ is used for optimization. A zero mask removes that response from the update; a one mask leaves its GRPO loss unchanged. A group-level selector uses the same decision for all responses from one prompt. With $L_{gk}$ denoting response length and $\ell^{\mathrm{GRPO}}_{gk\ell}$ the standard loss for token $\ell$, the selected loss is
\begin{equation}
    \mathcal{L}_{\mathrm{sel}}
    =\frac{1}{GK}\sum_{g=1}^{G}\sum_{k=1}^{K}
      \frac{m_{gk}}{L_{gk}}\sum_{\ell=1}^{L_{gk}}
      \ell^{\mathrm{GRPO}}_{gk\ell}.
    \label{eq:selection}
\end{equation}
The baseline sets every $m_{gk}=1$. Our selection configurations are \texttt{difffilter}, \texttt{maxvar}, \texttt{gfpo}, and \texttt{topk}.

\paragraph{Reweighting: change how strongly each token or response is learned.}
Reweighting keeps generated responses in the update but changes how strongly they contribute. Let $w_{gk\ell}\geq0$ be the weight on token $\ell$. A response-level method uses one weight for all tokens in a response; a token-level method can assign different weights within that response:
\begin{equation}
    \mathcal{L}_{\mathrm{rew}}
    =\frac{1}{GK}\sum_{g=1}^{G}\sum_{k=1}^{K}
      \frac{1}{L_{gk}}\sum_{\ell=1}^{L_{gk}}
      w_{gk\ell}\,\ell^{\mathrm{GRPO}}_{gk\ell}.
    \label{eq:reweighting}
\end{equation}
The baseline sets every $w_{gk\ell}=1$; our reweighting methods normalize weights to have mean one over the relevant batch units. The reweighting configurations are \texttt{ar}, \texttt{per}, \texttt{softmax}, and \texttt{diffband}.

\paragraph{Mixture adaptation: change which domains supply future prompts.}
Mixture methods act before the next rollout. Let $\mathcal{D}_d$ be the prompt distribution for domain $d$, and let $p_t(d)$ be its sampling probability at step $t$, with $p_t(d)\geq0$ and $\sum_{d=1}^{D}p_t(d)=1$. The GRPO objective under this training-data mixture is
\begin{equation}
    \mathcal{L}_{\mathrm{mix},t}(\theta)
    =\sum_{d=1}^{D}p_t(d)\,
      \mathbb{E}_{\substack{x\sim\mathcal{D}_d,\\
      y_{1:K}\sim\pi_{\theta_{\mathrm{old}}}(\cdot\mid x)}}
      \left[
      \frac{1}{K}\sum_{k=1}^{K}\frac{1}{L_k}
      \sum_{\ell=1}^{L_k}\ell^{\mathrm{GRPO}}_{k\ell}(\theta)
      \right].
    \label{eq:mixture}
\end{equation}
The fixed control uses $p_t(d)=1/D$ throughout training; with our three domains, this is $(1/3,1/3,1/3)$. A mixture method changes $p_t$ over time but leaves the per-response GRPO loss unchanged. Thus, selection and reweighting modify the current update after rollout, whereas mixture adaptation changes which prompts generate the next set of rollouts. The next subsection and Table~\ref{tab:methods} specify how each implemented mixture method computes $p_t$.

The timing also clarifies the compute comparison. Selection and reweighting occur after rollout, so they do not reduce generation cost. Selection is not renormalized and can reduce the number of tokens used in the update, whereas reweighting preserves a mean weight of one. Mixture adaptation leaves the per-response update unchanged and changes only the source of future prompts.

\subsection{Implemented methods and controlled comparisons}

The intervention family describes \emph{what} a method changes; the signal describes \emph{how} it chooses what to change. For example, a method may use the fraction of the five responses to a prompt that are correct, which we call the group solve rate $q_g$. Another method may score each response by its mean absolute GRPO advantage $a_{gk}$. Formally,
\begin{equation}
    a_{gk}=\frac{1}{L_{gk}}\sum_{\ell=1}^{L_{gk}}|A_{gk\ell}|,
    \qquad
    q_g=\frac{1}{K}\sum_{k=1}^{K}\mathbb{1}[r_{gk}>0.5],
    \label{eq:signals}
\end{equation}
where $A_{gk\ell}$ is the advantage of token $\ell$ and $r_{gk}$ is the verified reward. We also use the rollout policy's token probability $\pi_{gk\ell}$ and the efficiency score $r_{gk}/L_{gk}$. Mixture methods summarize each domain over a rolling 50-observation window using mean reward $\bar r_{t,d}$, mean absolute advantage $\bar a_{t,d}$, or reward slope $s_{t,d}$. Here $n_{t,d}$ is the number of prompts sampled from domain $d$ and $N_t=\sum_d n_{t,d}$ is the total.

This distinction gives us a direct comparison. The methods \texttt{per}, \texttt{softmax}, and \texttt{topk} all rank responses using $a_{gk}$. The first two convert that score into a continuous loss weight; \texttt{topk} uses it to keep or discard a response. Comparing them asks whether the same signal is more useful for reweighting or for selection.

Table~\ref{tab:methods} lists the complete set of released configurations. Several implement or closely adapt published proposals: \texttt{ar} follows Advantage Reweighting~\cite{yang2025lowprob}, \texttt{difffilter} is a stricter DAPO-style filter~\cite{yu2025dapo}, \texttt{maxvar} follows PODS~\cite{xu2025pods}, and \texttt{gfpo} uses GFPO's reward-per-token criterion~\cite{gfpo2025}. The mixture methods draw on DoReMi, DUMP, and Teacher--Student Curriculum Learning~\cite{xie2023doremi,wang2025dump,matiisen2017teacher}. The remaining configurations are controlled variants built from the same signals. For compactness, the table writes $i$ for a response indexed by a prompt--response pair $(g,k)$.

\begin{table}[t]
    \centering
    \small
    \begin{tabular}{>{\raggedright\arraybackslash}p{1.7cm}>{\raggedright\arraybackslash}p{1.7cm}>{\raggedright\arraybackslash}p{3.2cm}>{\raggedright\arraybackslash}p{7.0cm}>{\raggedright\arraybackslash}p{0.9cm}}
    \toprule
    \rowcolor{dcaiblue!6}Log Name & Family & Signal & Implemented Intervention & Level \\
    \midrule
    \rowcolor{dcaiblue!4}\multicolumn{5}{l}{\textit{Control}} \\
    \addlinespace[2pt]
    \texttt{baseline}    & baseline & none & $w_{i\ell}=1$ for every valid response token & control \\
    \midrule
    \multicolumn{5}{l}{\textit{Selection: fixed uniform domain mixture}} \\
    \addlinespace[2pt]
    \texttt{difffilter}  & selection & group solve rate $q_g$ & retain the whole group iff $0.2<q_g<0.8$; a stricter DAPO-style filter~\cite{yu2025dapo} & group \\
    \texttt{maxvar}      & selection & outcome rewards within group & retain $\operatorname{round}(0.5|g|)$ responses whose subset maximizes reward variance (PODS)~\cite{xu2025pods} & group \\
    \texttt{gfpo}        & selection & efficiency $r_i/L_i$ & retain the top three of five responses within each prompt group~\cite{gfpo2025} & group \\
    \texttt{topk}        & selection & mean $|A|$, $a_i$ & retain the highest-scoring $50\%$ of responses in the batch & response \\
    \midrule
    \rowcolor{dcaiblue!4}\multicolumn{5}{l}{\textit{Reweighting: fixed uniform domain mixture}} \\
    \addlinespace[2pt]
    \texttt{ar}          & reweighting & token probability $\pi_{i\ell}$ & $w_{i\ell}\propto0.5\pi_{i\ell}+0.5$; damp low-probability tokens~\cite{yang2025lowprob} & token \\
    \texttt{per}         & reweighting & mean $|A|$, $a_i$ & $w_i\propto(a_i+\epsilon)^{0.5}$; PER-inspired weighting of the current batch, not replay~\cite{schaul2016prioritized} & response \\
    \texttt{softmax}     & reweighting & mean $|A|$, $a_i$ & $w_i\propto\exp(a_i/T)$ with $T=1$ & response \\
    \texttt{diffband}    & reweighting & outcome reward $r_i$ & give $2\times$ weight to responses between the batch reward quartiles and $1\times$ otherwise, then normalize & response \\
    \midrule
    \rowcolor{dcaiblue!4}\multicolumn{5}{l}{\textit{Mixture adaptation: no within-domain selection or reweighting}} \\
    \addlinespace[2pt]
    \texttt{static}      & baseline & ignored & $p_t(d)=1/3$ & domain \\
    \texttt{reward\_gap} & mixture & rolling mean reward $\bar r_d$ & $p_t(d)\propto\exp((\max_{d'}\bar r_{d'}-\bar r_d)/T)$, with a $0.05$ floor; favors lagging domains & domain \\
    \texttt{dump\_ucb}   & mixture & mean $|A|$, $\bar a_d$, and count $n_d$ & $u_d=\bar a_d+c\sqrt{2\log(N+1)/(n_d+1)}$; $p_t(d)\propto\exp(u_d/T)$~\cite{wang2025dump} & domain \\
    \texttt{tscl}        & mixture & reward slope $s_d$ & $p_t(d)\propto\exp(|s_d|/T)$; favors domains with greater learning progress or forgetting~\cite{matiisen2017teacher} & domain \\
    \bottomrule
    \end{tabular}
    \caption{The 13 logged configurations, separated into selection, reweighting, and mixture adaptation. Reweighting formulas are mean-normalized, and all mixture probabilities use $T=1$ and a $0.05$ floor.}
    \label{tab:methods}
\end{table}

The canonical \texttt{dump\_ucb} and \texttt{tscl} implementations use the distinct signals shown in the table. Twelve earlier proxy runs that supplied reward levels to both methods are retained for provenance but excluded from every canonical mean.

\section{Experimental Setup}
\label{sec:pipeline}

Within each comparison, we keep the model, corpus, verifier, optimizer, rollout budget, and evaluation protocol fixed. This section describes the common setup used throughout the experiments.

The platform uses one shared driver for rollout, verification, and GRPO optimization. Each method is registered at one of the three intervention points in Figure~\ref{fig:overview} by specifying its signal, operator, and hyperparameters. A campaign generator expands model--method--seed combinations into runs, and the evaluators write all 12 benchmark scores to the same record format. The same records feed the released table-reconstruction and statistical-analysis scripts, so a new policy can change the intervention without changing the surrounding training and evaluation code.

\subsection{Data, training, and evaluation}
\label{sec:data}
\label{sec:harnesses}

The training corpus contains $15{,}000$ prompts, split equally among math, logic, and science. Math combines \texttt{math\_dapo}, DeepScaler, and GSM8K prompts with boxed-answer verification; logic uses procedurally generated Knights\&Knaves puzzles with an assignment checker; science uses SciQ multiple-choice questions with exact letter matching. The released records retain source and domain metadata for every prompt.

All campaigns use verl v0.5+ and GRPO with five rollouts per prompt, a KL coefficient of $10^{-3}$, and prompt and response limits of 1024 and 8192 tokens. Runs use $300$ optimizer steps and checkpoint every $100$ steps unless the campaign explicitly studies $1000$-step training. Comparisons use matched seeds; the breadth grid uses seeds $\{1,2,3\}$, while the Qwen-base primary matrix and higher-seed replications use seeds $\{1,\ldots,12\}$.

We evaluate every run on 12 benchmarks: five math sets (MATH-500, AIME-2024, OlympiadBench, MinervaMath, and GSM8K), four logic sets (Knights\&Knaves, two BBH tasks, and ZebraLogic), and three science sets (MMLU-Pro Chemistry, MMLU-Pro Physics, and GPQA-Diamond). Decoding is deterministic. Math and GPQA allow up to 8192 output tokens; the shared logic and MMLU-Pro evaluator uses 4096.

Training and evaluation formats can disagree without producing an obvious training failure. Before each campaign, we therefore run a calibration smoke test that checks boxed-answer and multiple-choice parsing on a reference checkpoint. We also audit the 15,000 training prompts against all evaluation items. The audit finds no exact or normalized matches and no 13-gram Jaccard similarity above $0.5$; the released artifact includes the audit outputs and reconstruction script.

\subsection{Run coverage}
\label{sec:coverage}

The experimental matrix contains 591 runs: the 171-run breadth study, 108 added runs that extend three baseline--selector comparisons from 3 to 12 seeds, 33 mechanism and strength controls, 108 additional Qwen2.5-7B-base runs, and 171 base-model scale and family runs from Revision Plan~6. The Qwen-base primary block completes all 13 configurations with 12 seeds each, including a campaign-local comparison of three adaptive mixtures with a fixed mixture.

The \textbf{Overall} score first averages within MATH, LOGIC, and SCIENCE, then averages the three domain scores, so each domain receives equal weight. Compact tables report Overall together with paired intervals; the Qwen-base primary table and the Llama replication table expose all 12 benchmark means. Header abbreviations are M500 (MATH-500), A24 (AIME24), Oly (OlympiadBench), Min (MinervaMath), K\&K (Knights\&Knaves), LD (BBH Logical Deduction), Track (BBH Object Tracking), Zebra (ZebraLogic), and Chem/Phys (MMLU-Pro Chemistry/Physics).

\section{Main Result on a Base Model with Training Headroom}
\label{sec:analysis}
\label{sec:f2}

Our main experiment uses Qwen2.5-7B-base, whose untrained checkpoint leaves substantial room for post-training. All 13 configurations are evaluated with 12 matched seeds. The comparison has two parts: selection and reweighting methods are measured against uniform sampling, while adaptive mixtures are measured against a fixed, equal mixture of math, logic, and science data.

\subsection{Training Headroom}

Before comparing data policies, we measure how much the shared GRPO recipe changes each representative starting model. Table~\ref{tab:qwen-base-headroom} reports the untrained checkpoint and the uniform-GRPO control under the same deterministic 12-benchmark evaluation. Both base-model rows use 12 matched seeds; the interval is shown for the Qwen2.5-7B-base comparison, which is the primary inferential setting.


\begin{center}
    \renewcommand{\arraystretch}{1.12}
    \begin{tabular}{lccccc}
    \toprule
    \rowcolor{dcaiblue!2}Base Model & Seeds & Untrained & Uniform GRPO & \textbf{$\Delta$} & 95\% CI \\
    \midrule
    Qwen2.5-7B-base      & 12 & 42.01 & 49.77 & \textcolor{red!75!black}{\textbf{+7.76}} & $[7.28,\ 8.25]$ \\
    Llama-3.1-8B-base    & 12 & 10.87 & 21.12 & \textcolor{red!75!black}{\textbf{+10.25}} & $[7.75,\ 12.33]$ \\
    \bottomrule
    \end{tabular}
    \captionof{table}{Training headroom in two representative base-model settings. Uniform GRPO is compared with the corresponding untrained checkpoint. Untrained scores are deterministic evaluations of fixed checkpoints; the Qwen2.5-7B-base interval reflects variation across its 12 trained baseline seeds.}
    \label{tab:qwen-base-headroom}
\end{center}

Uniform GRPO improves by $+7.76$ points on Qwen-base and $+10.25$ points on Llama-base. These gains make it unlikely that the null data-policy result is simply due to the model failing to learn. They also motivate the Qwen2.5-7B-base block as the primary test: it combines substantial headroom with complete method coverage and 12 matched seeds.

For a direct view of the primary experiment, Table~\ref{tab:qwen-base-12bench} reports the mean score on each benchmark after averaging the 12 matched seeds. The differences are not concentrated in one domain: methods that are relatively strong on LOGIC are not uniformly strong on MATH or SCIENCE. Overall therefore summarizes a genuinely multi-domain comparison rather than a single benchmark group.

\begin{table}[t]
    \centering
    \footnotesize
    \renewcommand{\arraystretch}{1.10}
    \setlength{\tabcolsep}{2.5pt}
    \begin{tabular}{lcccccccccccccc}
    \toprule
    & \multicolumn{5}{c}{\textcolor{dcaiblue}{MATH}} & \multicolumn{4}{c}{\textcolor{dcaiblue}{LOGIC}} & \multicolumn{3}{c}{\textcolor{dcaiblue}{SCIENCE}} & \multicolumn{2}{c}{\textcolor{dcaiblue}{SUMMARY}} \\
    \cmidrule(lr){2-6}\cmidrule(lr){7-10}\cmidrule(lr){11-13}\cmidrule(lr){14-15}
    \rowcolor{dcaiblue!5}\cellcolor{white}Method & M500 & A24 & Oly & Min & GSM8K & K\&K & LD & Track & Zebra & GPQA & Chem & Phys & \textbf{Overall} & 95\% CI \\
    \midrule
    baseline   & 84.45 & 14.44 & 38.06 & 34.93 & 91.42 & 20.46 & 65.77 & 78.75 & 29.78 & 33.71 & 53.17 & 57.05 & 49.77 & $[49.29,\ 50.26]$ \\
    difffilter & 84.23 & 15.01 & 37.78 & 35.06 & 91.10 & 22.38 & 65.76 & 76.97 & 32.28 & 32.58 & 53.75 & 56.42 & 49.85 & $[49.10,\ 50.61]$ \\
    maxvar     & 84.18 & 14.16 & 37.77 & 35.81 & 90.91 & 18.38 & 65.13 & 78.13 & 29.47 & 31.65 & 53.13 & 56.93 & 49.19 & $[48.57,\ 49.82]$ \\
    gfpo       & 83.78 & 13.88 & 37.88 & 34.94 & 91.31 & 17.62 & 64.80 & 76.16 & 30.53 & 31.73 & 53.23 & 56.03 & 48.88 & $[48.03,\ 49.73]$ \\
    topk       & 84.30 & 14.17 & 37.82 & 34.69 & 91.42 & 20.12 & 66.31 & 76.69 & 30.13 & 32.79 & 53.02 & 56.33 & 49.39 & $[48.67,\ 50.11]$ \\
    ar         & 84.63 & 14.45 & 37.30 & 35.24 & 91.18 & 19.79 & 66.23 & 77.21 & 31.55 & 33.29 & 52.63 & 55.78 & 49.50 & $[49.00,\ 49.99]$ \\
    per        & 81.93 & 12.51 & 36.59 & 33.48 & 90.11 & 22.83 & 66.10 & 73.18 & 30.82 & 33.54 & 52.53 & 56.72 & 48.92 & $[46.35,\ 51.48]$ \\
    softmax    & 83.77 & 13.88 & 37.28 & 35.36 & 90.92 & 21.12 & 66.04 & 76.62 & 30.93 & 33.04 & 53.95 & 56.13 & 49.54 & $[48.59,\ 50.50]$ \\
    diffband   & 84.13 & 15.83 & 37.75 & 35.54 & 91.07 & 18.79 & 67.23 & 77.33 & 30.07 & 33.80 & 53.73 & 56.58 & 49.75 & $[49.06,\ 50.45]$ \\
    static     & 84.60 & 15.00 & 37.51 & 35.02 & 91.27 & 19.00 & 66.04 & 75.11 & 29.05 & 33.16 & 52.48 & 55.43 & 49.00 & $[48.52,\ 49.48]$ \\
    reward\_gap & 83.95 & 11.38 & 37.66 & 33.85 & 91.37 & 21.25 & 66.36 & 77.21 & 31.58 & 32.66 & 53.35 & 56.87 & 49.46 & $[49.00,\ 49.91]$ \\
    dump\_ucb  & 84.67 & 13.62 & 37.79 & 34.60 & 91.21 & 19.42 & 66.79 & 78.02 & 31.25 & 31.90 & 53.22 & 57.67 & 49.61 & $[49.16,\ 50.07]$ \\
    tscl       & 84.13 & 14.72 & 37.68 & 34.19 & 91.27 & 16.54 & 66.08 & 78.29 & 30.10 & 32.24 & 53.47 & 56.27 & 49.16 & $[48.27,\ 50.04]$ \\
    \bottomrule
    \end{tabular}
    \caption{Twelve-benchmark means for the complete Qwen2.5-7B-base primary experiment. Each row averages 12 matched seeds; Overall gives equal weight to the MATH, LOGIC, and SCIENCE domain means. The final column is the 95\% $t$ interval for the seed-level Overall mean; paired method comparisons are reported in Table~\ref{tab:qwen-base-main}.}
    \label{tab:qwen-base-12bench}
\end{table}

\subsection{Selection and Reweighting Add No Reproducible Gain}

Table~\ref{tab:qwen-base-main} compares uniform sampling with all four selection and all four reweighting configurations, pairing each difference by training seed. This is the paper's primary method comparison: it combines complete method coverage with enough seeds to estimate the size of the remaining differences.


\begin{center}
    \renewcommand{\arraystretch}{1.10}
    \setlength{\tabcolsep}{4.5pt}
    \begin{tabular}{lcccc}
    \toprule
    \rowcolor{dcaiblue!3}Family & Method & Mean Score & Paired $\Delta$ & 95\% CI \\
    \midrule
    baseline & - & 49.77 & --- & --- \\
    \midrule
    \multirow{4}{*}{selection}
      & difffilter & 49.85 & \textcolor{red!75!black}{+0.08} & {$[-0.67,\ 0.82]$} \\
      & maxvar     & 49.19 & \textcolor{dcaiblue}{-0.58} & {$[-1.39,\ 0.23]$} \\
      & gfpo       & 48.88 & \textcolor{dcaiblue}{-0.90} & {$[-1.87,\ 0.08]$} \\
      & topk       & 49.39 & \textcolor{dcaiblue}{-0.38} & {$[-1.14,\ 0.38]$} \\
    \midrule
    \multirow{4}{*}{reweighting}
      & ar         & 49.50 & \textcolor{dcaiblue}{-0.28} & {$[-1.15,\ 0.60]$} \\
      & per        & 48.92 & \textcolor{dcaiblue}{-0.86} & {$[-3.28,\ 1.57]$} \\
      & softmax    & 49.54 & \textcolor{dcaiblue}{-0.23} & {$[-1.44,\ 0.98]$} \\
      & diffband   & 49.75 & \textcolor{dcaiblue}{-0.02} & {$[-0.61,\ 0.57]$} \\
    \bottomrule
    \end{tabular}
    \captionof{table}{Qwen2.5-7B-base selection and reweighting results with 12 matched seeds. Overall gives equal weight to MATH, LOGIC, and SCIENCE. Differences and paired $t$ intervals are relative to uniform sampling.}
    \label{tab:qwen-base-main}
\end{center}

None of the eight paired intervals excludes zero. Including the baseline, the nine policy means span $0.97$ point, about one eighth of the $7.76$-point gain from uniform GRPO. The selector ordering also changes when the comparison uses all 12 seeds: \texttt{topk} leads in the original three-seed subset, whereas \texttt{difffilter} has the highest 12-seed mean and \texttt{topk} falls below the baseline. The four reweighting estimates range from $-0.86$ to $-0.02$, and every interval crosses zero. At the precision of this experiment, changing the policy adds no reproducible gain, even though GRPO itself is clearly effective.

\subsection{Adaptive Mixtures Show No Reproducible Gain}

Table~\ref{tab:qwen-base-mixture} compares three adaptive mixtures with a \texttt{static} control. The checkpoint and training recipe are identical within this campaign; only the proportions of math, logic, and science prompts change during training. The \texttt{static} row is a campaign-local control, so its mean need not equal the primary uniform-baseline mean in Table~\ref{tab:qwen-base-main}; all mixture differences are paired within this campaign.


\begin{center}
    \renewcommand{\arraystretch}{1.10}
    \setlength{\tabcolsep}{4.5pt}
    \begin{tabular}{lcccc}
    \toprule
    \rowcolor{dcaiblue!3}Family & Method & Mean Score & Paired $\Delta$ & 95\% CI \\
    \midrule
    baseline & static & 49.00 & --- & --- \\
    \midrule
    \multirow{3}{*}{mixture}
      & reward\_gap & 49.46 & \textcolor{red!75!black}{+0.45} & $[-0.09,\ 1.00]$ \\
      & dump\_ucb   & 49.61 & \textcolor{red!75!black}{+0.61} & $[-0.02,\ 1.25]$ \\
      & tscl        & 49.16 & \textcolor{red!75!black}{+0.16} & $[-0.61,\ 0.93]$ \\
    \bottomrule
    \end{tabular}
    \captionof{table}{Qwen2.5-7B-base mixture-adaptation results with 12 matched seeds. Differences and paired $t$ intervals are relative to the fixed, equal \texttt{static} mixture. Every interval includes zero.}
    \label{tab:qwen-base-mixture}
\end{center}

All three adaptive mixtures have positive point estimates relative to the fixed mixture, but every paired interval includes zero and the means span only $0.61$ point. Training logs confirm that the methods changed the realized domain proportions. In this setting, changing the mixture changes which data the model sees, but it does not produce a reproducible improvement in final accuracy.

\section{Cross-Model and Evaluation Checks}
\label{sec:experiments}

Section~\ref{sec:analysis} establishes the main result on Qwen2.5-7B-base with 12 seeds and complete method coverage. We then examine three possible sources of variation: model family, model scale, and evaluation coverage. The first two checks use the corrected base-model extension. The last check recomputes the same runs under alternative benchmark summaries.

\subsection{Llama-3.1-8B-base Evaluation}

The corrected 12-seed Llama-3.1-8B-base evaluation uses the same protocol as the Qwen-base comparison. The untrained checkpoint scores $10.87$ Overall, while uniform GRPO reaches $21.12$, a gain of $10.25$ points (95\% CI $[7.75,12.33]$). Thus the common GRPO recipe is effective on Llama as well; the comparison is whether a data policy adds to that gain. Table~\ref{tab:llama12} gives the benchmark-level means.

\begin{center}
    \centering
    \footnotesize
    \renewcommand{\arraystretch}{1.10}
    \setlength{\tabcolsep}{2.5pt}
    \begin{tabular}{lcccccccccccccc}
    \toprule
    & \multicolumn{5}{c}{\textcolor{dcaiblue}{MATH}} & \multicolumn{4}{c}{\textcolor{dcaiblue}{LOGIC}} & \multicolumn{3}{c}{\textcolor{dcaiblue}{SCIENCE}} & \multicolumn{2}{c}{\textcolor{dcaiblue}{SUMMARY}} \\
    \cmidrule(lr){2-6}\cmidrule(lr){7-10}\cmidrule(lr){11-13}\cmidrule(lr){14-15}
    \rowcolor{dcaiblue!5}\cellcolor{white}Method & M500 & A24 & Oly & Min & GSM8K & K\&K & LD & Track & Zebra & GPQA & Chem & Phys & \textbf{Overall} & 95\% CI \\
    \midrule
    baseline   & 32.18 & 0.00 & 4.71 & 7.38 & 47.09 & 3.71 & 39.87 & 23.95 & 19.13 & 27.49 & 19.35 & 23.45 & 21.12 & $[20.01,\ 22.23]$ \\
    difffilter & 29.12 & 0.27 & 5.05 & 8.49 & 39.36 & 4.42 & 40.60 & 24.07 & 18.07 & 25.85 & 18.23 & 21.47 & 20.03 & $[18.46,\ 21.60]$ \\
    maxvar     & 32.98 & 0.55 & 4.85 & 8.12 & 44.87 & 3.92 & 38.31 & 23.57 & 16.52 & 25.42 & 19.65 & 22.07 & 20.41 & $[19.52,\ 21.30]$ \\
    topk       & 31.22 & 0.00 & 4.66 & 7.54 & 44.55 & 4.29 & 41.13 & 26.23 & 17.22 & 25.76 & 19.75 & 22.82 & 20.86 & $[19.78,\ 21.95]$ \\
    \midrule
    gfpo       & 31.23 & 0.27 & 4.92 & 7.94 & 46.77 & 4.21 & 38.99 & 27.89 & 18.55 & 27.53 & 19.77 & 22.93 & 21.35 & $[19.81,\ 22.89]$ \\
    ar         & 31.10 & 0.27 & 4.70 & 8.57 & 42.68 & 4.46 & 39.43 & 27.27 & 18.38 & 24.92 & 18.32 & 20.83 & 20.40 & $[18.99,\ 21.82]$ \\
    per        & 32.72 & 0.00 & 4.52 & 7.32 & 48.30 & 4.62 & 39.11 & 27.94 & 16.85 & 27.31 & 20.68 & 23.88 & 21.55 & $[20.59,\ 22.52]$ \\
    softmax    & 29.32 & 0.00 & 4.47 & 7.13 & 38.36 & 3.29 & 34.26 & 22.09 & 12.55 & 25.72 & 19.27 & 21.27 & 18.66 & $[16.91,\ 20.42]$ \\
    diffband   & 33.55 & 0.27 & 4.81 & 8.87 & 48.54 & 3.54 & 42.58 & 26.39 & 21.60 & 26.30 & 20.77 & 22.52 & 21.98 & $[20.76,\ 23.19]$ \\
    static     & 32.67 & 0.00 & 4.78 & 8.18 & 47.80 & 3.83 & 41.94 & 26.12 & 19.12 & 26.77 & 20.82 & 23.68 & 21.73 & $[20.71,\ 22.76]$ \\
    reward\_gap & 31.62 & 0.00 & 4.92 & 8.18 & 44.56 & 4.71 & 41.09 & 26.18 & 20.00 & 25.76 & 20.13 & 23.65 & 21.34 & $[19.99,\ 22.69]$ \\
    dump\_ucb  & 30.07 & 0.55 & 4.36 & 7.51 & 42.79 & 3.92 & 39.06 & 22.11 & 15.37 & 26.43 & 18.53 & 22.50 & 19.89 & $[18.00,\ 21.77]$ \\
    tscl       & 33.78 & 0.82 & 5.53 & 7.71 & 48.92 & 4.00 & 38.64 & 22.09 & 17.08 & 26.05 & 18.65 & 21.80 & 20.66 & $[19.66,\ 21.65]$ \\
    \bottomrule
    \end{tabular}
    \captionof{table}{Twelve-benchmark means for the Llama-3.1-8B-base evaluation. Each row averages 12 matched seeds; Overall gives equal weight to the MATH, LOGIC, and SCIENCE domain means. The final column is the 95\% $t$ interval for the seed-level Overall mean.}
    \label{tab:llama12}
\end{center}

Relative to the baseline, the paired Overall differences for the three original selectors are $-1.09$ for \texttt{difffilter} (95\% CI $[-2.84,0.65]$), $-0.71$ for \texttt{maxvar} ($[-2.18,0.76]$), and $-0.26$ for \texttt{topk} ($[-1.76,1.24]$). All three intervals include zero. The corrected extension places the nine additional methods on the same scale as the controls: their means range from $18.66$ (\texttt{softmax}) to $21.98$ (\texttt{diffband}), while the reusable baseline is $21.12$. The intervals in Table~\ref{tab:llama12} describe each method's seed-level mean; they are not paired method comparisons, and their overlap does not identify a universal winner. The earlier 14--16 point Llama values were produced by an evaluator that silently scored unsupported math tasks as zero and are excluded.

\subsection{Base-Model Scale and Family Coverage}
\label{sec:base-scale}

The corrected extension also covers several Qwen2.5 base-model sizes and Llama-3.2-3B. The auxiliary sizes use three matched seeds, whereas the Qwen 7B row is the 12-seed anchor from Section~\ref{sec:analysis}. Table~\ref{tab:qwen-base-scale} makes the comparison explicit by reporting each mean together with its difference from the appropriate control. This is a planned six-configuration sweep---baseline, \texttt{difffilter}, \texttt{maxvar}, \texttt{softmax}, \texttt{static}, and \texttt{tscl}---rather than a filtered presentation of a complete 13-method run. The other configurations (\texttt{gfpo}, \texttt{topk}, \texttt{ar}, \texttt{per}, \texttt{diffband}, \texttt{reward\_gap}, and \texttt{dump\_ucb}) were not run at these auxiliary sizes. The Qwen 14B row combines the available reusable control/selector runs with the new mixture-campaign runs. We use these rows to describe coverage and direction, not as standalone significance tests.

\begin{center}
    \setlength{\tabcolsep}{1.4pt}
    \begin{tabular}{l c c c >{\scriptsize}c c >{\scriptsize}c c >{\scriptsize}c c c >{\scriptsize}c}
    \toprule
    \multirow{2}{*}{Base Model} & \multirow{2}{*}{Seeds} & \multicolumn{1}{c}{Uniform} & \multicolumn{4}{c}{Selection} & \multicolumn{2}{c}{Reweighting} & \multicolumn{3}{c}{Mixture} \\
    \cmidrule(lr){3-3}\cmidrule(lr){4-7}\cmidrule(lr){8-9}\cmidrule(lr){10-12}
    & & baseline & difffilter & \multicolumn{1}{r}{$\Delta$U} & maxvar & \multicolumn{1}{r}{$\Delta$U} & softmax & \multicolumn{1}{r}{$\Delta$U} & static & tscl & \multicolumn{1}{r}{$\Delta$S} \\
    \midrule
    Qwen2.5-1.5B-base & 3 & 28.14 & 27.31 & \textcolor{dcaiblue}{-0.83} & 26.96 & \textcolor{dcaiblue}{-1.18} & 27.83 & \textcolor{dcaiblue}{-0.31} & 27.47 & 27.37 & \textcolor{dcaiblue}{-0.10} \\
    Qwen2.5-3B-base & 3 & 37.31 & 36.52 & \textcolor{dcaiblue}{-0.79} & 36.76 & \textcolor{dcaiblue}{-0.55} & 36.78 & \textcolor{dcaiblue}{-0.53} & 36.83 & 37.06 & \textcolor{red!75!black}{+0.23} \\
    Qwen2.5-7B-base & 12 & 49.77 & 49.85 & \textcolor{red!75!black}{+0.08} & 49.19 & \textcolor{dcaiblue}{-0.58} & 49.53 & \textcolor{dcaiblue}{-0.24} & 49.00 & 49.16 & \textcolor{red!75!black}{+0.16} \\
    Qwen2.5-14B-base & 3 & 59.38 & 60.58 & \textcolor{red!75!black}{+1.20} & 58.54 & \textcolor{dcaiblue}{-0.84} & 59.14 & \textcolor{dcaiblue}{-0.24} & 58.22 & 59.65 & \textcolor{red!75!black}{+1.43} \\
    Llama-3.2-3B-base & 3 & 13.16 & 12.80 & \textcolor{dcaiblue}{-0.36} & 11.47 & \textcolor{dcaiblue}{-1.69} & 13.45 & \textcolor{red!75!black}{+0.29} & 12.74 & 12.96 & \textcolor{red!75!black}{+0.22} \\
    \bottomrule
    \end{tabular}
    \captionof{table}{Overall means in the corrected base-model scale and family sweep. Each cell for a selector or reweighting method reports its mean followed by the difference from the uniform baseline; the \texttt{static} column is the fixed-mixture control, and the \texttt{tscl} difference is relative to that control. Auxiliary rows use three seeds, while Qwen2.5-7B-base uses the 12-seed anchor. \textcolor{red!75!black}{Red} indicates an increase and \textcolor{dcaiblue}{blue} indicates a decrease.}
    \label{tab:qwen-base-scale}
\end{center}

The signs vary across models rather than following a common pattern. For example, \texttt{difffilter} is below the uniform baseline at 1.5B and 3B but slightly above it at 7B and 14B, while \texttt{maxvar} is below the baseline in every row. The mixture comparison also changes direction: \texttt{tscl} is below its fixed control at 1.5B but above it at the larger Qwen sizes and on Llama-3.2-3B. These three-seed rows therefore show why a single cross-model winner is not evident; the complete inferential comparison remains the Qwen2.5-7B-base anchor.

\subsection{Method Rankings Depend on Evaluation-Domain Coverage}
\label{sec:threedomain}

We ask whether the leading method remains the same when the evaluation covers all three training domains. It does not: a math-heavy six-benchmark summary (five math sets plus GPQA-Diamond, with no logic benchmark) and the domain-balanced 12-benchmark summary select different leading methods, whereas alternative summaries retaining all 12 benchmarks largely agree.

The reason is that the methods exhibit domain trade-offs. On Qwen2.5-7B-Instruct, for example, \texttt{diffband} has the highest LOGIC mean ($59.0$) and the lowest SCIENCE mean ($44.2$). An evaluation that omits LOGIC therefore measures a materially different profile rather than merely averaging fewer benchmarks.

Table~\ref{tab:aggregation-sensitivity} makes this effect explicit for the same nine Qwen2.5-7B-Instruct configurations. We call the math-heavy summary Math-Heavy-6 below; it averages five math benchmarks and GPQA-Diamond and contains no logic benchmark. DB-12 first averages within MATH, LOGIC, and SCIENCE and gives the three domains equal weight. Macro-12 weights benchmarks equally, while Item-12 weights them by evaluation-set size.

\begin{table}[ht]
    \centering
    \begin{tabular}{lcc}
    \toprule
    \rowcolor{dcaiblue!3}Evaluation Summary & Rank $\rho$ vs DB-12 & Method Spread \\
    \midrule
    DB-12 (domain-balanced) & $1.00$ & $3.31$ \\
    Macro-12              & $0.88$ & $2.79$ \\
    Item-12               & $0.88$ & $2.17$ \\
    Math-Heavy-6          & $-0.33$ & $0.90$ \\
    \bottomrule
    \end{tabular}
    \caption{Sensitivity of the nine Qwen2.5-7B-Instruct configuration means to evaluation-domain coverage and aggregation. Rank correlation is Spearman's $\rho$ relative to DB-12; spread is the highest minus lowest configuration mean in percentage points. Each configuration mean uses the same three trained seeds.}
    \label{tab:aggregation-sensitivity}
\end{table}

Math-Heavy-6 ranks \texttt{topk} and \texttt{diffband} first, whereas DB-12 ranks \texttt{gfpo} and \texttt{difffilter} first. The rankings are negatively correlated ($\rho=-0.33$), and the between-method spread changes from $0.90$ to $3.31$ points. By contrast, Macro-12 and Item-12 both correlate strongly with DB-12 ($\rho=0.88$), and leave-one-out checks range from $0.78$ to $0.98$ (Appendix~\ref{app:aggregation}). The disagreement is therefore driven mainly by the omitted logic domain, not by small changes in weighting once all 12 benchmarks are retained.

These checks leave the primary conclusion unchanged. The corrected Llama evaluation places the additional configurations on the same score scale as the original controls and finds no common winner in their observed means. The base-model scale and family sweep likewise shows no monotone advantage for any method. The aggregation analysis adds a separate caution: if a domain is omitted, the apparent winner can change even when the underlying runs are identical.

\section{Conclusion}
\label{sec:conclusion}

\textsc{DataFlex-RL} evaluates whether existing RLVR data policies add to the gains delivered by a shared GRPO recipe. On Qwen2.5-7B-base, uniform GRPO improves Overall accuracy by $7.76$ points. None of the eight selection or reweighting methods has a paired interval excluding zero relative to uniform sampling, and none of the three adaptive mixtures has one relative to a fixed equal mixture; their means span $0.97$ and $0.61$ point, respectively. On Llama-3.1-8B-base, GRPO improves by $10.25$ points (95\% CI $[7.75,12.33]$); the original selector intervals include zero, and the corrected extension places nine additional methods on the same score scale without producing a clear common winner in their means.

The cross-model and evaluation checks show why these comparisons require both adequate training headroom and careful evaluation. The corrected base-model extension does not reveal a common winner, and benchmark coverage can change the apparent winner even when summaries retaining all 12 benchmarks largely agree. The evidence therefore supports a scoped conclusion: under the controlled recipes tested here, the evaluated selection, reweighting, and mixture policies do not provide a reproducible improvement over uniform training at the achieved precision. It does not establish equivalence or rule out gains under longer training, other hyperparameters, or different policy families. The released records, logs, configurations, and reconstruction scripts make these checks auditable and reusable for future policies.

\begingroup
\small
\setlength{\bibsep}{2pt}
\bibliographystyle{plainnat}
\bibliography{main}
\endgroup

\clearpage

\beginappendix


\section{Additional Analyses}

\subsection{Sensitivity to the Evaluation Summary}
\label{app:aggregation}

DB-12 is the mean of the MATH, LOGIC, and SCIENCE domain scores. We compare it with three alternatives on the nine-configuration Qwen2.5-7B-Instruct campaign. Math-Heavy-6 averages five math benchmarks with GPQA-Diamond and omits logic; Macro-12 gives every benchmark equal weight; Item-12 weights benchmarks by evaluation-set size. Math-Heavy-6 and DB-12 have negatively correlated method rankings (Spearman $\rho=-0.33$, Kendall $\tau=-0.28$), and the range of method means changes from $0.90$ to $3.31$ points. Macro-12 and Item-12 both correlate with DB-12 at $\rho=0.88$. Leave-one-benchmark-out correlations range from $0.78$ to $0.98$. The ranking change is therefore driven primarily by the omitted logic domain rather than by the weighting used once all 12 benchmarks are retained.

\subsection{Selector Diagnostics}
\label{app:selector-diagnostics}

Table~\ref{tab:mechanism-controls} compares each original selector with controls for retention count, update volume, and intervention strength. Matched random keeps the same number of groups or responses but chooses them randomly. Matched update tokens continues training until the run has used at least as many nonzero update tokens as the 300-step baseline. Alternative strengths widen the filtering band or change the keep fraction.

\begin{center}
    \centering
    \begin{tabular}{lcccc}
    \toprule
    Selector & Targeted & Matched Random & Matched Update Tokens & Alternative Strength(s) \\
    \midrule
    difffilter & $53.81\pm1.38$ & $54.29\pm1.85$ & $54.33\pm4.09$ & $54.48\pm0.90$ \\
    maxvar    & $53.42\pm0.56$ & $50.93\pm3.17$ & $54.64\pm1.67$ & $54.30\pm1.33$ / $52.34\pm0.79$ \\
    topk      & $52.86\pm0.66$ & $52.29\pm1.54$ & $51.58\pm0.11$ & $53.42\pm1.63$ / $53.44\pm1.25$ \\
    \bottomrule
    \end{tabular}
    \captionof{table}{Three-seed selector diagnostics on Qwen2.5-7B-Instruct (mean $\pm$ sample standard deviation). Alternative strengths use the full nonzero solve-rate band for \texttt{difffilter} and keep fractions $0.25/0.75$ for \texttt{maxvar} and \texttt{topk}.}
    \label{tab:mechanism-controls}
\end{center}

The controls move the three selectors in different directions. Targeted selection exceeds matched random for \texttt{maxvar} and \texttt{topk}, but not for \texttt{difffilter}; matching update tokens raises \texttt{difffilter} and \texttt{maxvar}, but lowers \texttt{topk}. No single targeting, update-volume, or strength effect explains all three methods. In a separate response-efficiency analysis, \texttt{gfpo} and the baseline have nearly identical observed accuracy on five math benchmarks ($72.21\%$ and $72.19\%$), while \texttt{gfpo} reduces mean response length from 1568 to 1497 characters and raises correct answers per thousand characters from $0.461$ to $0.482$.

\end{document}